\documentclass[sigconf]{acmart}
 \acmSubmissionID{278}

\usepackage{booktabs}

\newcommand{\bL}{\mathcal{L}}

\newcommand{\win}{\omega_\mathrm{i}}
\newcommand{\wout}{\omega_\mathrm{o}}

\usepackage{mathtools}
\usepackage{xspace}

\makeatletter
\newcommand\paragraphNew{\@startsection{paragraph}{4}{\parindent}%
  {-.5\baselineskip \@plus -2\p@ \@minus -.2\p@}%
  {-3.5\p@}%
  {\ACM@NRadjust{\@parfont}}}
	\makeatother
	
\AtBeginDocument{%
  \providecommand\BibTeX{{%
    \normalfont B\kern-0.5em{\scshape i\kern-0.25em b}\kern-0.8em\TeX}}}

\newcommand{\revise}[1]{\textcolor{black}{#1}}

\copyrightyear{2024}
\acmYear{2024}
\setcopyright{acmlicensed}\acmConference[SIGGRAPH Conference Papers '24]{Special Interest Group on Computer Graphics and Interactive Techniques Conference Conference Papers '24}{July 27-August 1, 2024}{Denver, CO, USA}
\acmBooktitle{Special Interest Group on Computer Graphics and Interactive Techniques Conference Conference Papers '24 (SIGGRAPH Conference Papers '24), July 27-August 1, 2024, Denver, CO, USA}
\acmDOI{10.1145/3641519.3657410}
\acmISBN{979-8-4007-0525-0/24/07}

\begin{document}


\title{Woven Fabric Capture with a Reflection-Transmission Photo Pair}

\author{Yingjie Tang}
\orcid{0009-0002-5633-0079}
\affiliation{
    \institution{Nankai University}
    \city{Tianjin}
    \country{China}
}
\email{lxtyin.ac@gmail.com}

\author{Zixuan Li}
\orcid{0009-0004-2424-9529}
\affiliation{
    \institution{Nankai University}
    \city{Tianjin}
    \country{China}
}
\email{zixuan.li_2001@outlook.com}

\author{Milo\v{s} Ha\v{s}an}
\orcid{0000-0003-3808-6092}
\affiliation{
    \institution{Adobe Research}
    \city{San Jose}
    \country{USA}
}
\email{milos.hasan@gmail.com}

\author{Jian Yang}
\orcid{0000-0003-4800-832X}
\affiliation{
    \institution{Nanjing University of Science and Technology}
    \city{Nanjing}
    \country{China}
}
\email{csjyang@njust.edu.cn}

\author{Beibei Wang}
\orcid{0000-0001-8943-8364}
\authornote{Corresponding author.}
\affiliation{
    \institution{School of Intelligence Science and Technology, Nanjing University}
    \city{Suzhou}
    \country{China}
}
\email{beibei.wang@nju.edu.cn}






\begin{abstract}
Digitizing woven fabrics would be valuable for many applications, from digital humans to interior design. Previous work introduces a lightweight woven fabric acquisition approach by capturing a single reflection image and estimating the fabric parameters with a differentiable geometric and shading model. The renderings of the estimated fabric parameters can closely match the photo; however, the captured reflection image is insufficient to fully characterize the fabric sample reflectance. For instance, fabrics with different thicknesses might have similar reflection images but lead to significantly different transmission. We propose to recover the woven fabric parameters from \emph{two} captured images:  reflection and transmission. At the core of our method is a differentiable bidirectional scattering distribution function (BSDF) model, handling reflection and transmission, including single and multiple scattering. We propose a two-layer model, where the single scattering uses an SGGX phase function as in previous work, and multiple scattering uses a new azimuthally-invariant microflake definition, which we term ASGGX. This new fabric BSDF model closely matches real woven fabrics in both reflection and transmission. We use a simple setup for capturing reflection and transmission photos with a cell phone camera and two point lights, and estimate the fabric parameters via a lightweight network, together with a differentiable optimization. We also model the out-of-focus effects explicitly with a simple solution to match the thin-lens camera better. As a result, the renderings of the estimated parameters can agree with the input images on both reflection and transmission for the first time. The code for this paper is at \href{https://github.com/lxtyin/FabricBTDF-Recovery}{https://github.com/lxtyin/FabricBTDF-Recovery.}



\end{abstract}


\begin{teaserfigure}
\centering
\includegraphics[width=\textwidth]{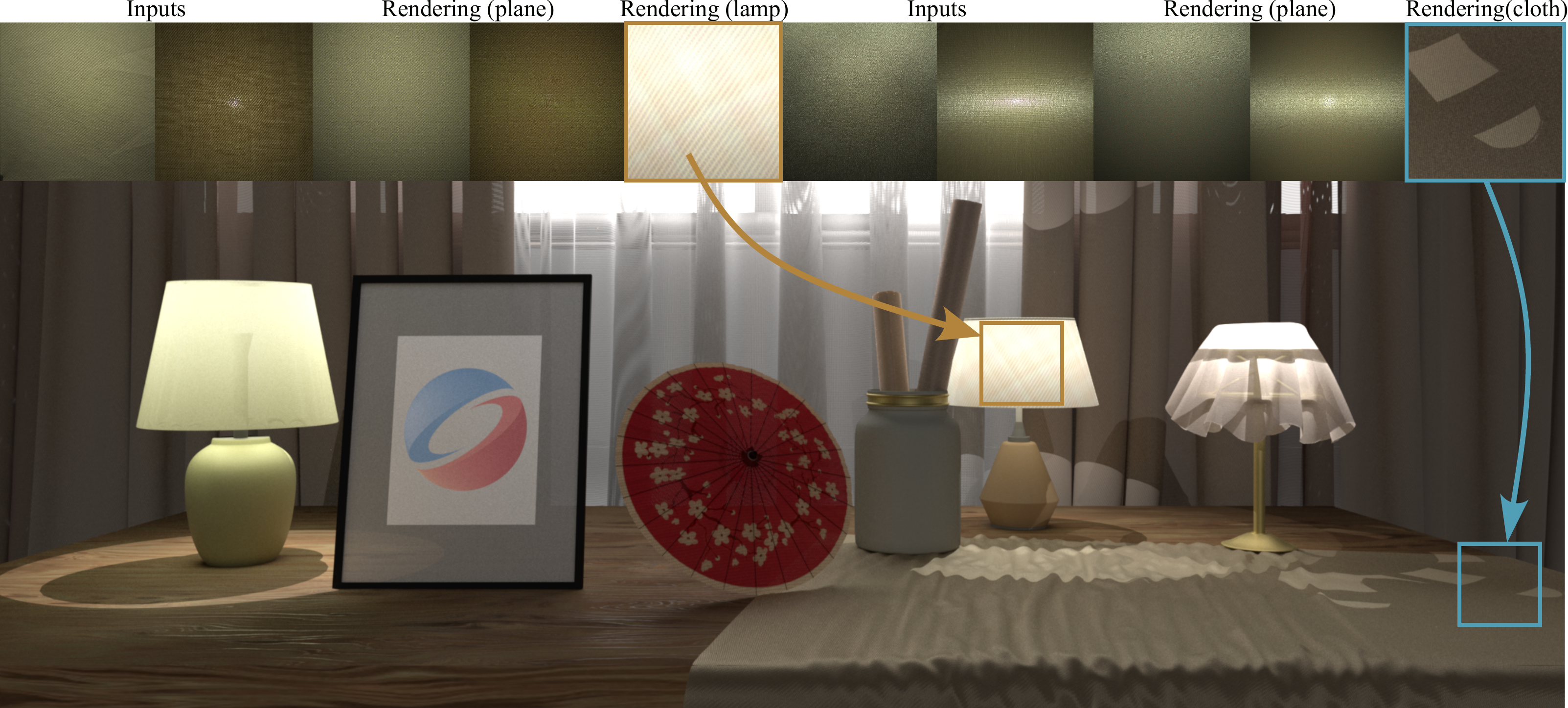}
\caption{Given two photos of a woven fabric sample (front-lit and back-lit), our approach estimates the parameters of our proposed woven fabric material model. Re-rendered results with estimated parameters closely match the input photos on the top. The resulting fabric parameters can be used in a rendered scene, either directly or after further editing, \revise{by using spatially-varying diffuse color maps.} }
\label{fig:teaser}
\end{teaserfigure}

\maketitle

\section{Introduction}
\label{sec:intro}

Rendering fabrics is valuable for many applications, such as interior visualization, fabric design, virtual reality, digital humans, etc. However, creating high-quality digital fabric assets requires extensive work, even for experienced artists. The alternative is to capture the fabric materials from the real world. Historically, capture required specialized devices and complex pipelines; recently, lightweight fabric capture has become an active research topic. In this paper, we focus on lightweight capture of woven fabrics: a common and important subset.

Recent work by Jin et al.~\shortcite{Jin:2022:inverse} introduced a lightweight woven fabric capture approach by taking a single photo in a simple setup. At the core of their method is a differentiable procedural  model and a bidirectional reflectance distribution function (BRDF) specialized for woven fabrics, used for parameter recovery through inverse rendering. The recovered parameters can match the captured photos very well, in terms of both highlights and structure. However, their method only models the reflection and recovers the parameters under the supervision of only the reflection image. Fabrics with different thicknesses might have similar reflection images, but very different backlighting (light transmission) behavior. Transmission is of critical importance if the fabric is used as a window or lamp shade, but can also be important when used as clothing. This effect can no longer be estimated from a single image; both the reflection and transmission images are needed to recover the complete set of fabric parameters.

In this paper, we propose a simple configuration to capture two images of a flat fabric sample from the same camera but with two light positions, front and back, for lightweight parameter recovery. Our method recovers the parameters of woven fabrics by a combination of a small neural network and differentiable optimization, \revise{following a reconstruction pipeline similar to the work of Jin et al. ~\shortcite{Jin:2022:inverse}}, matching the captured reflection and transmission images. Fabrics with the recovered parameters can be used in a final rendering engine.

A key component of our method is a new bidirectional scattering distribution function (BSDF) model for woven fabrics, modeling both single and multiple scattering for both reflection and transmission. Previous works models the transmission of fabrics using the SpongeCake model~\cite{Wang:2021:Sponge, Zhu:2023:cloth}, which consists of volumetric layers with fiber-like microflake phase functions. They represent multiple scattering with a single scattering lobe but with modified parameters. Unfortunately, these models in their current form cannot fully explain the behavior of multiple scattering from woven fabrics in the transmission image, as shown in Fig.~\ref{fig:renderedMulti}. To this end, we extend SpongeCake to a new two-layer BSDF model for woven fabrics, whose novel component uses a specialized \revise{empirical} phase function for aggregated microflakes to model the multiple scattering from fibers. Furthermore, we also model gaps between yarns and the out-of-focus appearance of the lightsource to better match the transmission photographs. As a consequence, our recovered parameters can faithfully match the captured images on both synthetic and real data for several typical woven fabric patterns.
To summarize, our main contributions include:
\begin{itemize}
    \item a new phase function for aggregated microflakes to better characterize the appearance of multiple scattering from fibers,
    \item a new two-layer BSDF model with several key components to match real woven fabric reflection and transmission,
    \item a lightweight capture configuration that only requires two photos of a fabric sample under front and back point illumination.
\end{itemize}

\section{Related Work}
\label{sec:related}

\paragraph{Fabric models}

We briefly review woven fabric models, consisting of geometry and appearance models. 
The geometry of woven fabrics can be represented in different ways, including volume, curve, and surface-based models. The volume representation usually couples with the microflake model~\cite{Jakob:2010:microflake, heitz2015SGGX} to define fiber-like participating media. Curve-based fabric models usually use bidirectional curve scattering distribution functions (BCSDFs)~\cite{Marschner:2003:HairBCSDF, Chiang:2015:fur, Montazeri:2020:ply, Zhu:2023:yarn} to define the optical properties of each fiber or ply. Surface models rely on macroscopic fabric geometry, together with a BSDF as the surface shading model~\cite{IrawanAndMarschner2012, Sadeghi:2013:Cloth, Jin:2022:inverse, Zhu:2023:cloth}.  

Among these three groups, the volume and curve-based models can bring high realism at extreme close-up views at the cost of memory and computation. In contrast, surface-based models are lightweight and can achieve high-fidelity results at the macroscopic scale. We focus on the latter type of model in our paper.

\paragraph{Surface fabric models.}
Surface fabric models act as BRDFs (e.g., \cite{Adabala:2003:cloth} \cite{IrawanAndMarschner2012} \cite{Sadeghi:2013:Cloth}). They model the fabric structures with normals and tangents, which are used in the reflectance model. Recently, Jin et al.~\shortcite{Jin:2022:inverse} propose a lightweight procedural geometric and reflectance model based on the SpongeCake model~\cite{Wang:2021:Sponge}. Their method is simple and differentiable, allowing for inverse rendering. Zhu et al.~\shortcite{Zhu:2023:cloth} use the SpongeCake forward model to enable shadowing-masking from the yarns at the cost of a more complex formulation, and do not consider capture. 


\paragraph{Single-image fabric recovery.}
Several methods have been proposed to recover fabrics with a single image as input at different levels. Schr{\"o}der et al.~\shortcite{schroder2015imagebased} and Wu et al.~\shortcite{Wu2019modeling} can achieve fiber-level detail, but rely on manual selection of model parameters or require expensive time cost. Guarnera et al.~\shortcite{Guarnera2017wfmc} estimate the yarn parameters in the spatial and frequency domain at the yarn level at the cost of a complex pipeline. More related work can be found in the survey by Castillo et al.~\shortcite{Castillo2019recent}. Unlike the above work, our method targets a simple setup and lightweight pipeline. \citet{rodriguez2019automatic} aims at recovering the macroscopic color pattern textures for woven fabrics rather than the fabric parameters, which is out of our scope. 

A closely related work to ours is by Jin et al.~\shortcite{Jin:2022:inverse}, which estimates woven fabric parameters from a single photo with differentiable rendering. However, it only considers the reflection image, which we will show to be insufficient; our method utilizes both the reflection and transmission images for fabric parameter recovery.

\paragraph{Procedural material parameter estimation.}
Besides the above works specialized for fabrics, some approaches have been proposed for predicting general procedural material parameters by learning the mapping from the input image to the parameters with a neural network\revise{~\cite{hu2019novel, Shi:2020:MATch, garces2023towards, rodriguez2023umat}} or Bayesian framework~\cite{Guo:2020:Bayesian}. These methods are designed for general materials and are not optimal for fabrics, but our method shares some common components with these approaches, such as neural parameter initialization and differentiable optimization with perceptual losses.

\begin{table}[!t]
	\renewcommand{\arraystretch}{1.1}
 \label{tab:all_symbols}
	\caption{\label{tab:all_symbols} Parameters in our BSDF model. The top three parameters affect the yarn geometry, and the rest affect reflectance. The * indicates that this variable allows different values for warp and weft yarns.    }
\begin{small}
  \begin{tabular}{|l|c|l|}\hline
        $s$ & yarn size \\
		$\beta $ *& heightfield scaling factor \\ 
        $\xi$ *& gap scaling
        \\\hline
        
        $k_\mathrm{s}^\mathrm{s}, k_\mathrm{s}^\mathrm{m}$ *&specular albedo for single / multiple  \\
		$k_\mathrm{d}^\mathrm{r}, k_\mathrm{d}^\mathrm{t}$ &diffuse albedo for reflection / transmission \\
		$\alpha^\mathrm{s}, \alpha^\mathrm{m} $ *& roughness for single / multiple \\
		$\psi $ *& fiber twist angle \\
		$u_{\mathrm{max}} $ & maximum inclination angle \\
		$T^\mathrm{s}, T^\mathrm{m}$ *& thickness of the fabric for single / multiple\\
		$w$ & weight for the Lambertian term blending \\
		$U_\mathrm{s}(\xi)$ & randomness on the specular term \\
		$U_\mathrm{n}(\xi)$ & randomness on the normal and orientation \\
		$\mathrm{Q}$ & normal / orientation randomness level \\
        \hline
		\end{tabular}

\end{small}
\end{table}

\section{Background and motivation}
\label{sec:back}

Woven fabrics are manufactured by interweaving warp and weft yarns. Here we define the the warp as vertical and the weft as horizontal, as shown in Fig.~\ref{fig:two_layer_model}. We focus on several typical weave patterns, though the system is easily extensible to other patterns. 

\paragraph{Background on reflection capture}
Jin et al.~\shortcite{Jin:2022:inverse} propose a geometric and appearance model for woven fabrics as a spatially-varying BRDF defined on a macroscopic fabric surface, instead of using volumes~\cite{Zhao:2011:fabric} or curves~\cite{Zhu:2023:yarn}. They model yarn geometry as smooth bent cylinders, which do not have to be explicitly constructed, and instead yield normal vectors, tangents and other information used in their reflectance model. The reflectance model includes a specular and a diffuse term; the former is based on the SpongeCake model~\cite{Wang:2021:Sponge} with a fiber-like microflake phase function, and the latter is a double diffuse term considering both the macro-surface normal and the yarn normal. Their model can represent accurate woven fabric reflection, both in terms of highlight shapes and spatial texture. They recover the fabric parameters from a single captured reflection image with a neural parameter prediction network followed by optimization via differentiable rendering.

\paragraph{Motivation for adding transmission capture}
Recovering the fabric parameters solely from reflection cannot provide enough information to reproduce all fabric parameters, most obviously the thickness, which is not sensitive to reflection but strongly affects transmission. Most other parameters also affect transmission and can be estimated more accurately by matching the transmission image as well. Our solution is to use reflection and transmission together for fabric parameter recovery.

\section{Fabric bidirectional scattering distribution function}
\label{sec:forward}


Jin et al.~\shortcite{Jin:2022:inverse} provide an appearance model for woven fabric reflection. Their model is based on SpongeCake~\cite{Wang:2021:Sponge}, which supports transmission automatically. The straightforward way is to simply enable transmission in Jin et al.'s model. Unfortunately, we find that using the transmission as-is cannot accurately match multiple scattering in the transmission image, as shown in Fig.~\ref{fig:renderedMulti}. The problem turns out to be that the approximation of reusing the single-scattering lobe based on SGGX microflakes for multiple scattering, while generally reasonable, produces the largest error where the incoming and outgoing directions are pointing opposite each other, which happens to be in the center of the back-lit transmission image. Therefore, a key problem is to design a better multiple-scattering lobe, which can match real fabric transmission photographs closely. We propose a new BSDF (Sec.~\ref{sec:model}) to model fabric reflection and transmission, whose main new component is a better representation for the multiple scattering of microflake media (Sec.~\ref{sec:multiple}).

\subsection{An azimuthally-invariant phase function for multiple scattering}
\label{sec:multiple}

\begin{figure}[tb]
\centering
\includegraphics[width = 0.9\linewidth]{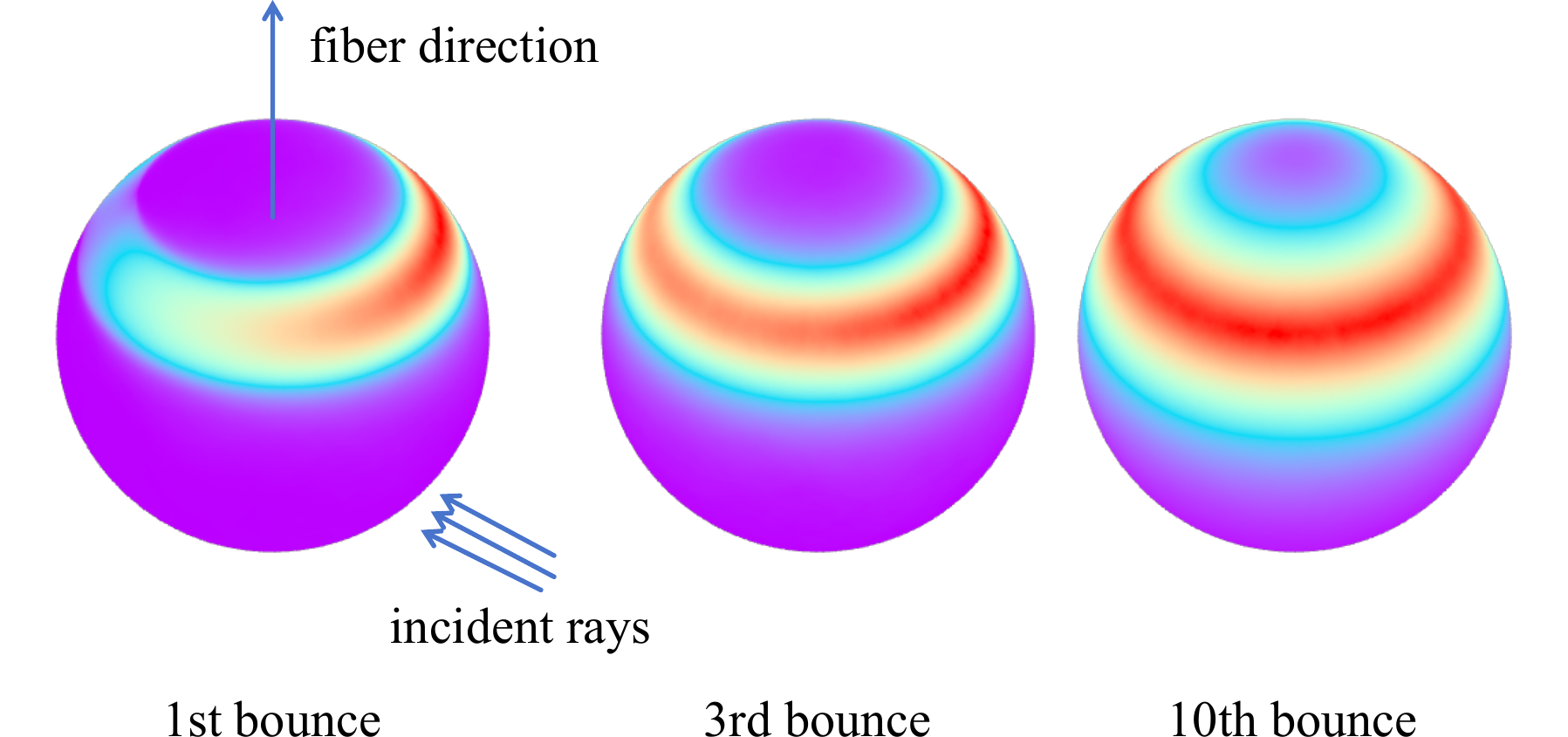}
\caption{ Given an incoming ray, we simulate the scattering among the fibers with different bounces using the SGGX phase function and visualize the distribution of the outgoing ray. The outgoing distribution becomes uniform along the azimuth angle. Representing such an azimuthally invariant distribution is beyond the capability of the SGGX phase function.}
\label{fig:mulbounce}
\end{figure}

\begin{figure}[tb]
\centering
\includegraphics[width = 0.9\linewidth]{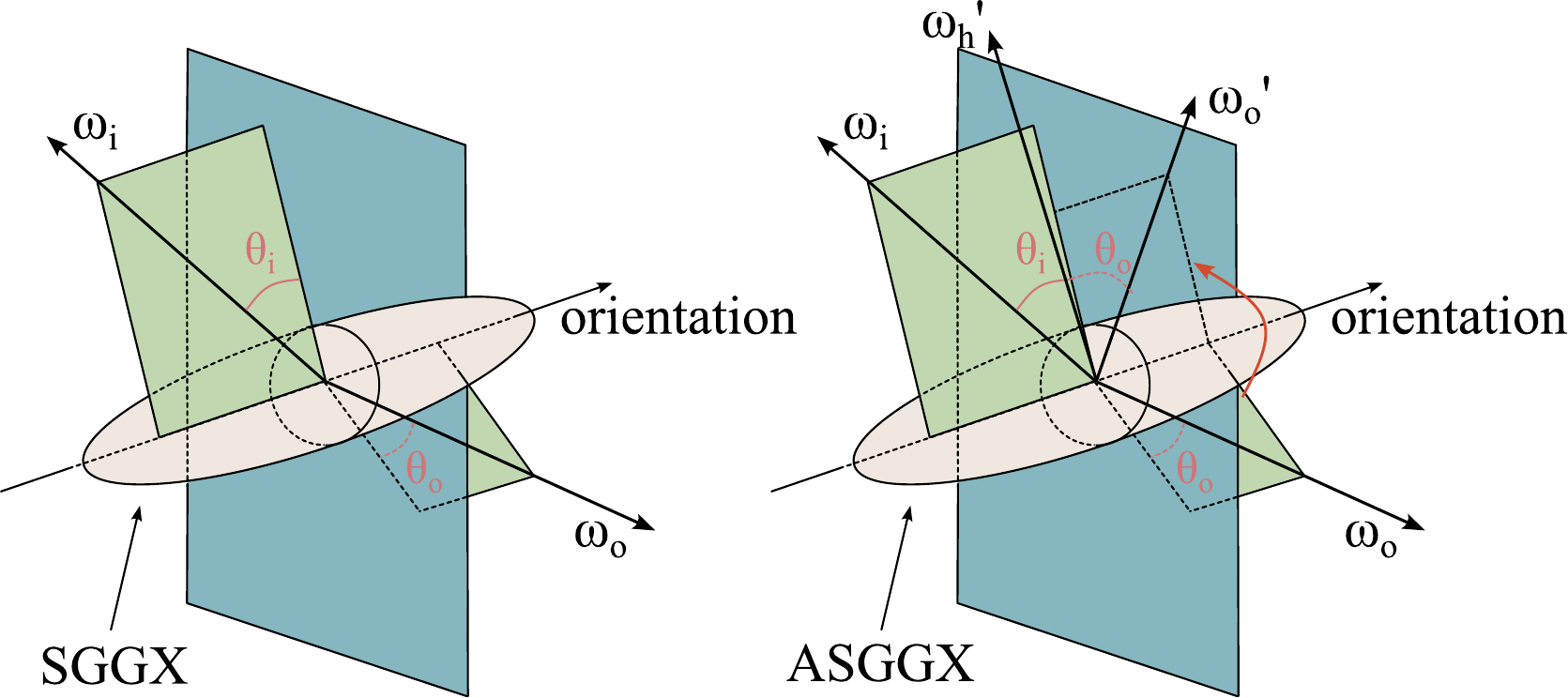}
\caption{ Configurations of the SGGX (left) and our ASGGX (right). In our ASGGX, $\omega_o$ is rotated to the same longitudinal plane as $\omega_i$, leading to a new vector $\omega_o'$. Then, $\omega_i$ and $\omega_o'$ form a half-vector $\omega_h'$, which is used to look up the microflake density.}
\label{fig:aggflake}
\end{figure}



\begin{figure}[tb]
\centering
\includegraphics[width = \linewidth]{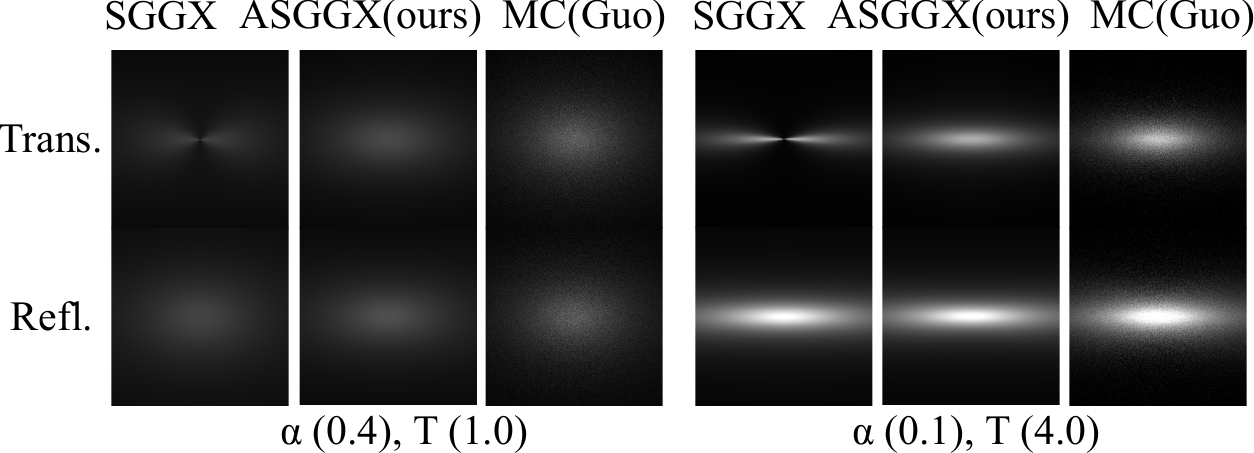}
\caption{
\revise{Multiple scattering comparison among SGGX, ASGGX (ours), and \revise{Monte-Carlo simulation \cite{Guo:2018:Layered}, which serves as a synthetic reference} on two different sets of parameters (roughness $\alpha$ and thickness $T$). Our method can closely match the \revise{reference} on both reflection and transmission, while the results by SGGX match the GT on reflection only.}
}
\label{fig:renderedMulti}
\end{figure}




As shown in Fig.~\ref{fig:mulbounce}, the multiple scattering distribution quickly becomes azimuthally uniform due to the diffusion of the multiple bounces among the microflakes. This behavior cannot be achieved by a single-scattering SpongeCake lobe using standard SGGX microflakes, no matter how the parameters are set. Hence, we design a specialized azimuthally-invariant microflake phase function, which we term ASGGX, to represent the multiple scattering distribution. This new phase function can be plugged into the SpongeCake  formulation to derive the corresponding BSDF.


Intuitively, we can think of an aggregation of microflakes as a single scattering event. Note that the overall fiber direction is given (as in standard SGGX), and we will assume a coordinate frame aligned with this direction. We propose an azimuthally-invariant phase function as follows: given an incoming direction $\win$ and outgoing direction $\wout$, we rotate them into the same arbitrary longitudinal plane, so that their azimuth angles become identical. We define the new phase function using the half-vector $\omega_h'$ computed from the \emph{modified} directions. For this definition to be valid, the longitudinal plane chosen above should not matter for the final result, which will be true in our case, because the microflake density is rotationally invariant around the fiber direction. For convenience we choose the plane that contains $\win$ for Fig.~\ref{fig:aggflake} and the discussion below.

More precisely, our azimuthally-invariant phase function $f_\mathrm{p}(\win \rightarrow \wout)$ has very similar formulation to SGGX, and reuses its density $D(\omega)$, but the half vector is computed differently:
\begin{equation}
     f_\mathrm{p}(\win \rightarrow \wout) = \frac{D(\omega_h')}{2\sigma(\win)}.
\end{equation}
Here, $\omega_h'$ is the half-vector between the modified (rotated) $\wout'$ and $\win$, as shown in Fig.~\ref{fig:aggflake}. The 2 in the denominator comes from the transformation $d\wout'=2|\wout' \cdot \omega_h'|d\omega_h'$, where $\wout'$ and $\omega_h'$ are treated as 2D unit vectors restricted to the longitudinal plane. This is different from the standard half-vector reflection Jacobian, $d\wout=4|\wout \cdot \omega_h|d\omega_h$. The functions $D$ and $\sigma$ are the same as for SGGX \cite{heitz2015SGGX}, and are rotationally invariant around the fiber direction.
The proposed azimuthally-invariant phase function satisfies both energy conservation and reciprocity. 


After establishing the phase function, we use the single scattering of this \emph{aggregated microflake} to represent the multiple scattering of the original microflakes. For that, we need to modify the relative thickness of the aggregated flake. The final multiple scattering is computed by considering the attenuation:
\begin{equation}
    f_m(\win, \wout) = \frac{k_s^mD_m(\omega_{h'})G_m(\win, \wout)}{2cos(\win)cos(\wout)}
    \label{eq:multiple}
\end{equation}
where $D_m$ and $G_m$ is specific for multiple scattering, detailed in the supplementary. They are the same form as SpongeCake but using modified thickness and roughness.

In Fig.~\ref{fig:renderedMulti}, we compare the rendered results of our model, the SpongeCake model and the reference which is rendered with Guo et al.~\shortcite{Guo:2018:Layered}. Here, the parameters are found by optimization.


\subsection{Two-layer microflake model}
\label{sec:model}


\begin{figure}[tb]
\centering
\includegraphics[width = 0.8\linewidth]{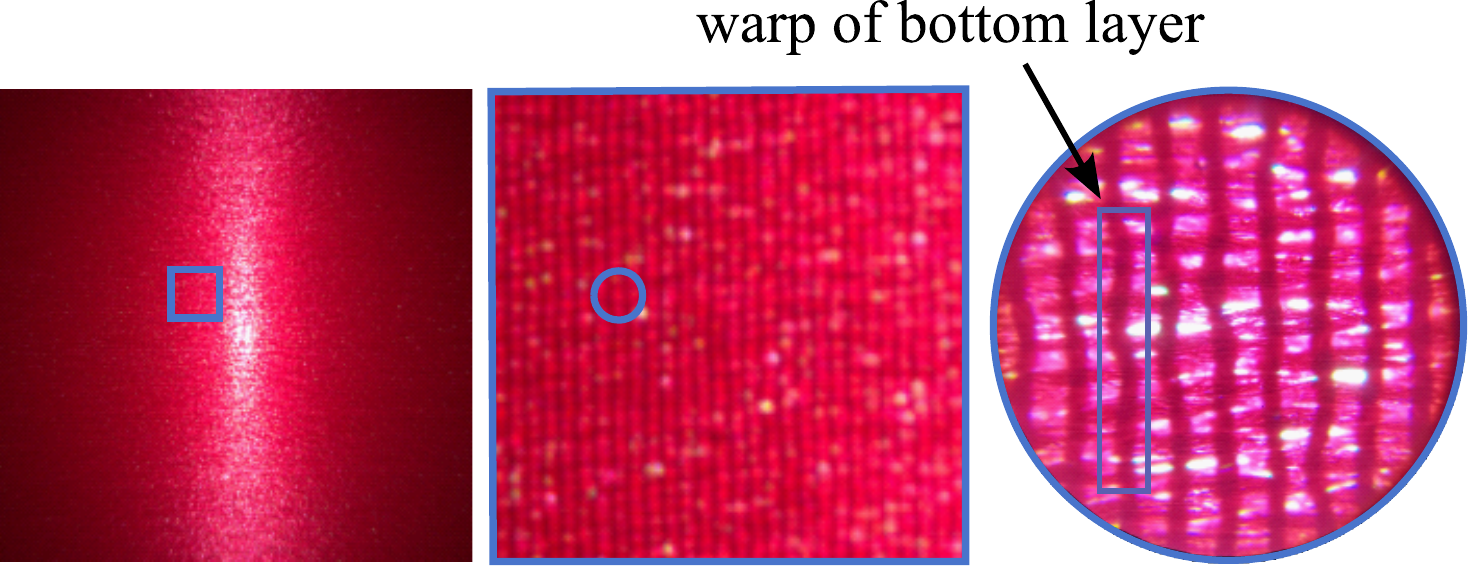}
\caption{Macro appearance and its zoom-in. Yarns of the bottom layer are also visible, which have a significant effect on the macro appearance.}
\label{fig:obs1}
\end{figure}

\begin{figure}[tb]
\centering
\includegraphics[width = 0.7\linewidth]{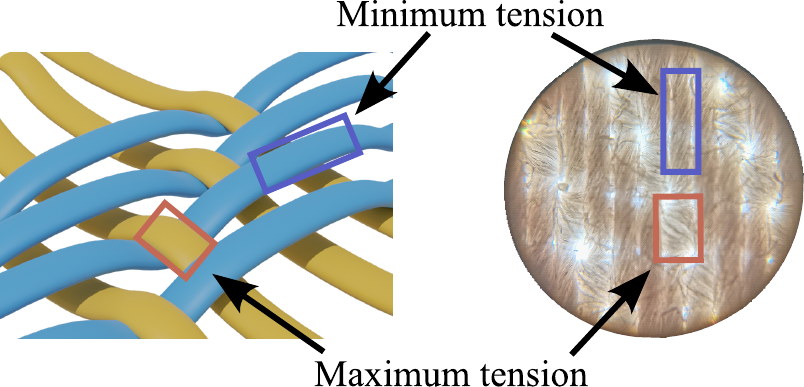}
\caption{The yarn becomes thinner at the intersection of two yarns because of the increased tension.}
\label{fig:obs2}
\end{figure}


Now we can model single scattering by a SpongeCake layer using the original SGGX microflake and the multiple scattering term by a layer using the new ASGGX microflake, which we observe to match the fabric photos better. However, we have several additional observations  (Fig.~\ref{fig:obs1}) from the captured transmission image, which will inform our final BSDF model.

First, in a transmission image, both the weft and warp yarns have an effect at a single pixel, which cannot be represented by the single-layer model that was sufficient in the work of \citet{Jin:2022:inverse} and \citet{IrawanAndMarschner2012}. Second, yarn thickness varies due to different tension at different points within a weave pattern. As the thickness decides the fraction of the light passing through, it significantly impacts the transmission, while it could be ignored in previous work considering only reflection. These observations are not modeled in previous work~\cite{Jin:2022:inverse, Zhu:2023:cloth}. Therefore, we introduce a two-layer yarn model, and propose a tension-aware thickness modulation function. 


\begin{figure}[tb]
\centering
\includegraphics[width = 0.8\linewidth]{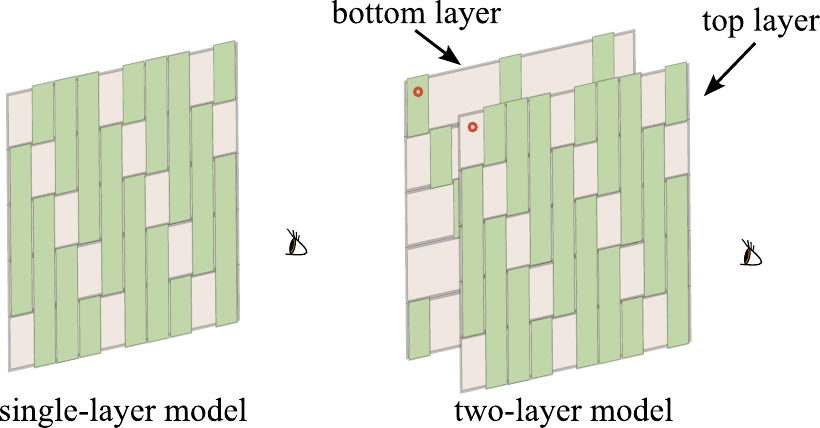}
\caption{Instead of using a single-layer model (left), we introduce a two-layer model (right), where both the weft and warp yarns are considered at each location (red dot).}
\label{fig:two_layer_model}
\end{figure}

\paragraph{Two-layer model}
We use two SpongeCake layers to represent, at each point on the fabric, the yarn on the top (closer to camera) and at the bottom (away from camera) respectively, as shown in Fig.~\ref{fig:two_layer_model}. In this model, each layer represents a yarn and its parameters depend on whether it is a weft or warp. Each of them has their own yarn parameter set, including diffuse albedo and roughness. All the parameters are summarized in Table.~\ref{tab:all_symbols}.

\paragraph{Tension-aware thickness function}
To characterize the appearance from the varying thickness, we propose a thickness modulation function for each yarn, depending on the location on the yarn. The thickness is defined by considering the scaling factor:
\begin{equation}
    T = T\times (S_\mathrm{min} + (1 - \mu) \times(1 - S_\mathrm{min})),
    \label{eq:thickness}
\end{equation}
where $S_\mathrm{min}$ is the minimum thickness scale, set as 0.5 for satin and twill, and 1.0 for plain in practice. $\mu$ is the tension level, defined as 0.0 at the center of a yarn's long part, 1.0 at the center of the yarn's compressed part, and linearly interpolated between the two.


\subsection{Final BSDF model}

We now define our full bidirectional scattering distribution function (BSDF), which includes three terms: single scattering term for two layers using the SGGX microflake phase function, low-order multiple scattering term for two layers using the new ASGGX phase function, and an additional modified diffuse term for high order scattering. As for the diffuse term, we use a blended formulation similar to \citet{Jin:2022:inverse}, considering both the macro surface normal and the micro-yarn normals on both sides. Each term is defined for both reflection and transmission. 

For a given surface location on the fabric, let $\win$ and $\wout$ be the incident (light) and outgoing (camera) directions in world space. Let $\omega_\mathrm{m}$ be the smooth macroscopic surface normal (e.g. interpolated from vertex normals). Our fabric shading model has three terms:
\begin{equation}
     f(\win, \wout) = f_\mathrm{s}(\win, \wout) + f_{\mathrm{m}}(\win, \wout) + f_{\mathrm{d}}^{\mathrm{r,t}}(\win, \wout).
     \label{eq:final}
\end{equation}
Here $f_\mathrm{s}(\win, \wout)$ represents single scattering and is a two-layer SpongeCake lobe using SGGX, $f_{\mathrm{m}}(\win, \wout)$ represents lower-order multiple scattering and is a two-layer SpongeCake lobe using ASGGX, and the diffuse term represents higher-order scattering:
\begin{equation}
     f_\mathrm{d}^{\mathrm{r}}(\win, \wout) = w\frac{k_\mathrm{d}^\mathrm{r} \left \langle\win \cdot \omega_\mathrm{n1} \right \rangle}{\pi \left \langle\win \cdot \omega_\mathrm{m} \right \rangle} + (1-w)\frac{k_\mathrm{d}^\mathrm{r} }{\pi},
\end{equation}
\begin{equation}
     f_\mathrm{d}^{\mathrm{t}}(\win, \wout) = w\frac{k_\mathrm{d}^\mathrm{t} 
    \left \langle\win \cdot \omega_\mathrm{n1} \right \rangle \cdot 
     \left \langle\win \cdot \omega_\mathrm{n2} \right \rangle
     }{\pi 
     \left \langle\win \cdot \omega_\mathrm{m} \right \rangle} 
     + (1-w)\frac{k_\mathrm{d}^\mathrm{t} }{\pi},
\end{equation}
where $\omega_{\mathrm{n1}}, \omega_{\mathrm{n2}}$ are the yarn normal of first layer and second layer respectively, and $k_d^r, k_d^r$ denote the diffuse albedo for the reflection and transmission, respectively. For reflection, we only consider the normal at the first layer, while for transmission, we use a product of cosine terms, since we empirically observe that both normals affect the diffuse component.  

Our single scattering term $f_\mathrm{r}^{\mathrm{s}}(\win, \wout)$ is the same as the single scattering of a two-layer SpongeCake model with the SGGX phase function~\cite{Wang:2021:Sponge}, including both reflection and transmission, where the thickness is scaled by the thickness scaling function from Eqn.~(\ref{eq:thickness}). The detailed formulation is shown in the supplementary. Finally, our multiple scattering term is also formulated as a two-layer SpongeCake model with the ASGGX phase function and its own parameters, as shown in Eqn.~(\ref{eq:multiple}). 

To summarize, our woven fabric shading model consists of the following parameters (see Table~\ref{tab:all_symbols}): a discrete weave pattern, single/multiple scattering albedo for weft/warp, diffuse albedo for reflection and transmission, roughness/thickness for weft/warp and single/multiple scattering respectively, yarn size for weft/warp, a height field scaling factor for weft/warp, a gap scaling factor for weft/warp (shown in Fig. 1 in the supplementary), a twist angle for weft/warp, blending weight of the diffuse term, randomness on the specular term, and a noise level performed on the height field scaling factor to control the intensity of the orientation / normal map randomness.

\section{Fabric parameter estimation}
\label{sec:estimation}

Based on the proposed fabric BSDF, we estimate the woven fabric parameters. We first introduce a simple setup for fabric sample capture (Sec.~\ref{sec:capturing}), and then introduce the pipeline for the parameter estimation (Sec.~\ref{sec:network} and ~\ref{sec:opt}).

\subsection{Measurement setup for fabrics}
\label{sec:capturing}


\begin{figure}[tb]
\centering
\includegraphics[width = \linewidth]{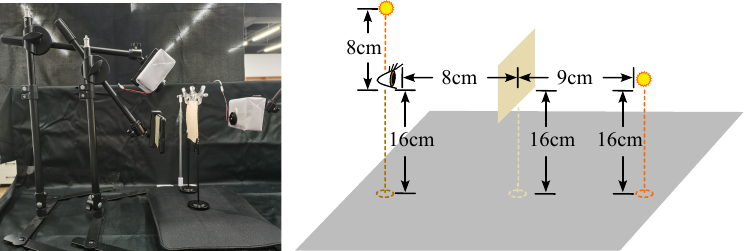}
\caption{The configuration to measure the real fabric data. We use one cell phone as camera and two small light sources to capture the reflection and transmission image respectively. The fabric sample is fixed by a holder. We also measure the distances in this configuration to reconstruct the same setup in synthetic renderings.}
\label{fig:measureConfig}
\end{figure}

We propose to capture a reflection-transmission photo pair for fabric parameter reconstruction, as shown in Fig.~\ref{fig:measureConfig}. We use one cell phone as camera and two point lights for illumination and put the fabric sample in-between two light sources with a holder. The captured raw images have a $4K$ resolution, and we crop and downsample them to a resolution of $512 \times 512$. We calibrate the light brightness and apply falloff due to lens vignetting, similar to Jin et al.~\shortcite{Jin:2022:inverse}.




\subsection{Neural network for fabric parameter prediction}
\label{sec:network}


\begin{figure}[tb]
\centering
\includegraphics[width = \linewidth]{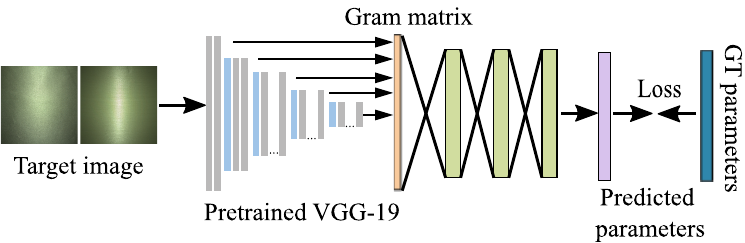}
\caption{Our network architecture.}
\label{fig:network}
\end{figure}


\begin{figure}[tb]
\centering
\includegraphics[width = \linewidth]{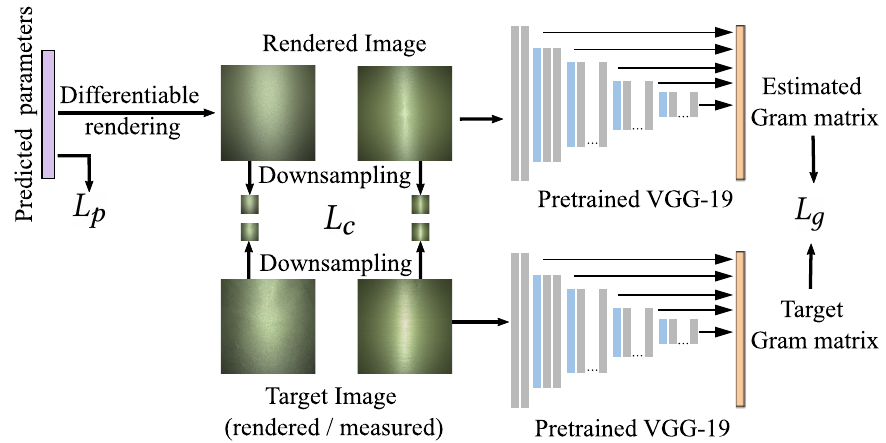}
\caption{Our optimization pipeline.}
\label{fig:optimization}
\end{figure}


\paragraph{Architecture.} Given the input reflection-transmission pair, we use a simple neural network (Fig.~\ref{fig:network}) to predict approximated parameter vectors \revise{and the pattern type}. We feed the two images into a pretrained VGG-19 network separately, and compute Gram matrices, resulting in two features, each a vector of size 610304. We concatenate the features from the two images and feed them into a fully connected (FC) module, which includes three intermediate layers (256 nodes per layer) with LeakyReLU activation function. The final FC layer outputs the predicted parameters (34 channels for our forward model). Note that this network is similar to the one by Jin et al.~\shortcite{Jin:2022:inverse}, except we use two images, one more layer, and different output channels.

\paragraph{Dataset generation.} We generate a rendered dataset of five weave patterns (twill, satin, plain, and 90-degree rotations of the twill and satin), The twist angle is set at -30 degrees for the twill and no twist for other patterns. We generate 1,280 images for each pattern with our shading model (Sec.~\ref{sec:forward}), by sampling the fabric parameter space, detailed in Table~$S2$ (Table~2 in the supplementary material). 

\paragraph{Training.} The loss function for network training is the $\bL_1$ difference between the ground truth parameters and the predicted parameters from the network. Our network is implemented in the PyTorch framework with the Adam solver, where the learning rate is set as 0.0001 and the batch size is set as 32. Only FC weights are updated during training (VGG / Gram matrix computation is frozen). Training took four hours on a single A40 GPU.
\subsection{Optimization with differentiable rendering }
\label{sec:opt}

We use the network-predicted fabric parameters as an initialization and perform optimization (Fig.~\ref{fig:optimization}) with differentiable rendering, which further improve the match, e.g. in color.

We render the reflection and transmission images in a differentiable PyTorch function. To better match the out-of-focus light appearance in transmission images caused by the camera focusing on the fabric plane, we project the point light to the rendered image, and generate a Gaussian around the projected center. The gaps between the yarns look up this Gaussian, simulating direct view of the light source. This way, our rendered result can match an out-of-focus effect, despite using a pin-hole camera model in the simulation. In practice, we set the Gaussian scaling as 8 and the variance as 20 pixels, but this depends on our specific camera and light, and could be calibrated for more generality.


We compute the difference between the rendered images and the target images to drive the optimization. We use several losses to measure the difference between the target images and the rendered images, including a VGG-19 Gram matrix loss $L_g$, a prior loss $L_p$ on the scaling factor $\beta$ and gap scaling factor $\xi$ for optimization robustness, and a pixel loss $L_c$ between down-sampled images with resolution $16\times16$ to improve color. 


%
Our final loss is defined as
\begin{eqnarray}
    L_{\mathrm{opt}} &= &L_g + w_1 L_{p} + w_2 L_c,\\
    L_{g} &= &\bL_1(\mathrm{Gram}(I), \mathrm{Gram}(R)), \\
    L_{p} &= &-\mathrm{log}\left(\mathrm{exp}\left(-\frac{(\beta - \mu_{\beta})^2} 
    {2\sigma_\beta^2}\right)\right)
    -\mathrm{log}\left(\mathrm{exp}\left(-\frac{(\xi - \mu_{\xi})^2} 
    {2\sigma_\xi^2}\right)\right),\quad \\
    L_{c} &= &\bL_1(I_{\mathrm{down}},R_{\mathrm{down}}).
    \label{eq:loss}
\end{eqnarray}
where $w_1$ and $w_2$ are set as 0.001 and 0.1 respectively, $\mu_{\beta}$ and $\sigma_{\beta}$ are the mean and the variance of the Gaussian prior on the scaling factor $\beta$, respectively. They are set as (1.0, 0.5) for the twill, (0.1, 0.5) for the satin and (1.0, 1.0) for the plain. The Gaussian prior on gap scaling factor $\xi$ is set as (0.9, 0.05) for satin and twill, (0.75, 0.1) for plain. We also optimize several discrete parameters, including the yarn density of the weft and warp, the twist angle and gap scaling, detailed in the supplementary. We use the Adam optimizer with learning rate 0.01 for 300 iterations, which takes about four minutes on an NVIDIA 4060 GPU.

\paragraph{Discussion}
\revise{Note that the tension-aware thickness is not optimized and treated as a fixed per-pattern property, as it is determined by the proportion of the yarn's long and compressed parts.}

\revise{During the optimization, we perform differentiable rendering with Eqn. (\ref{eq:final}), without considering the yarn-level shadowing-masking term ~\cite{Zhu:2023:cloth}, as the shadowing-masking effect in the current capture configuration is not obvious, and its formulation is not differentiable. Instead, we apply this term for the final mesh renderings (Figs.~\ref{fig:real} and ~\ref{fig:synthetic}) after estimating the parameters.}

\section{Results}
\label{sec:results}








We first show the results of our procedural parameter estimation model on both synthetic data and real data. Afterwards, we perform ablation studies on several key components. 
 
\subsection{Results of our inverse model}

\paragraph{Synthetic data.}
In Fig.~\ref{fig:synthetic}, we validate our method on synthetic data using five kinds of fabrics. Given two images as inputs, our network first predicts coarse parameters whose renderings roughly match the inputs but with some color bias and highlight mismatches. Then, the differentiable optimization addresses the above issues, producing a closer match between input images and renderings, which is also confirmed by the rendering of the draped cloth mesh. \revise{For further validation, we provide the estimated parameters (roughness and thickness) in Table 3 (supplementary), showing that the difference between the predicted and the ground-truth parameters is acceptable.}




\paragraph{Real data.}
In Fig.~\ref{fig:real}, we perform parameter estimation on real measured data and compare our method with Jin et al.~\shortcite{Jin:2022:inverse}. Since no ground truth parameters exist for the measured data, we compare the visual match between the input and rendered images with the estimated parameters. The renderings with the draped cloth mesh also show a plausible appearance for both reflection and transmission. For Jin et al.~\shortcite{Jin:2022:inverse}, we capture their input image in their suggested setup, wrapping the fabric around a cylinder. Their method can reproduce  reflection results matching the real photo. In theory, these estimated parameters can be used to render transmission as well. However, we find that their transmission prediction differs significantly from the captured photo due to the missing of some critical parameters (e.g., thickness) and the sub-optimal handling of multiple scattering in the original SpongeCake model. \revise{More results are shown in Fig. 2 (supplementary).}

\revise{We further validate our method by capturing the fabric samples from a novel view (by rotating them) and rendering them under the same configuration. The rendering of estimated parameters can match the real fabrics at the novel view, as shown in Fig. 3 (supplementary).}


We apply the estimated parameters from real data into a complex scene and further edit the parameters with spatially varying diffuse color maps to demonstrate the various appearances \revise{in Fig.~\ref{fig:teaser}}.


\subsection{Ablation studies}

\begin{figure}[tb]
\centering
\includegraphics[width = 1.0\linewidth]{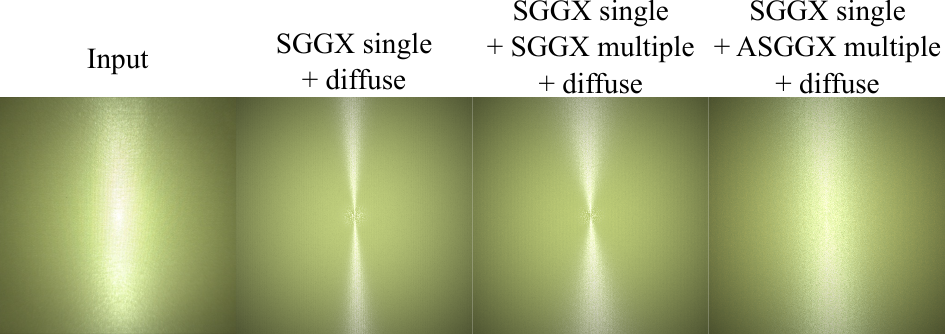}
\caption{ The impact of our ASGGX on fabric recovery, by comparing different shading models on a satin fabric. Our ASGGX outperforms the others and can match the highlights in the input transmission image. }
\label{fig:ablation_asggx}
\end{figure}



\begin{figure}[tb]
\centering
\includegraphics[width = 0.8\linewidth]{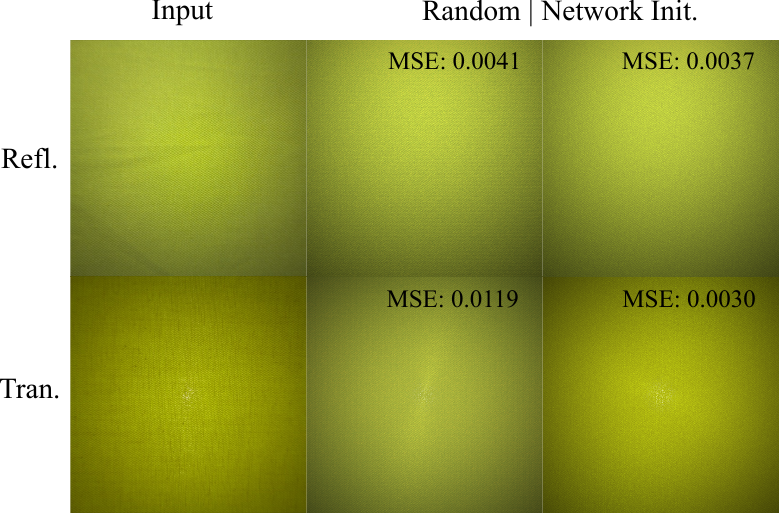}
\caption{Comparison between a random initialization and our network initialization, where the latter shows a lower error.}
\label{fig:ablation_network}
\end{figure}



\paragraph{Impact of the ASGGX phase function.}
Our ASGGX for multiple scattering is critical for transmission representation and recovery. We validate its impact by comparing fabric recovery with three different transmission shading models: 1) a diffuse term, 2) SGGX + diffuse, and 3) ASGGX + diffuse, where all of them have another SGGX for single scattering. We use the network for prediction in these three results and then perform the optimization with these shading models under the same settings (iterations and learning rate). As shown in Fig.~\ref{fig:ablation_asggx}, a single diffuse, or together with the SGGX, mismatch the highlights with the input images due to the characteristics of SGGX, while our solution (ASGGX + diffuse) produces a better match with the input. 


\paragraph{Impact of the two-layer model.}
Fabrics have vertical continuous yarns in the real capture, particularly for the satin, as shown in Fig.~\ref{fig:obs1}. Our two-layer model captures this appearance. To show its influence, we compare the renderings of estimated parameters optimized by our two-layer and single-layer models on the satin example in Fig.~\ref{fig:ablation_twolayer}. The single layer cannot produce the vertical continuous yarns, as the weft yarns always cut the warp yarns. In contrast, in the two-layer model, the light can pass through the weft and reach the warp even if the weft yarn is on the top, leading to a  continuous vertical yarn. 

\paragraph{Impact of the tension-aware thickness. }
We validate the influence of tension-aware thickness in Fig.~\ref{fig:ablation_thickscale}, by comparing with a constant thickness. By comparison, we find that the overall diagonal structure cannot be captured with a constant thickness due to the overlap between the two layers. This issue is addressed by our tension-aware thickness, which enhances this overall diagonal structure, leading to a better match with the real data.

\paragraph{Impact of the network prediction.}
\revise{Similar to Jin et al.~\shortcite{Jin:2022:inverse}, our method uses a network for initialization.} We show the impact of the network in Fig.~\ref{fig:ablation_network} by comparing the results with and without using the network for initialization. The results with network initialization show higher quality than a random initialization.

\paragraph{\revise{Impact of the loss function.}} \revise{We use three loss terms (the pixel loss, the Gram matrix loss, and the prior loss) in the optimization step. We validate their influence in Fig. 4 (supplementary). By comparison, the pixel loss and the Gram matrix loss reduce color bias, while the prior loss improves robustness. More detailed discussions are shown in the supplementary material.}

\subsection{Discussion and limitations}


\paragraph{Missing yarn variations and global features.} \revise{Our shading model does not consider the variations in yarn diameter, yarn sliding, or global features (e.g., wrinkles and flyaway fibers), leading to some mismatches in the estimated results. Our forward model can be extended to support all these features by introducing more complex procedural spatial variation; However, estimating many more parameters controlling these features may make the optimization more difficult with only two images as inputs.}

\revise{\paragraph{Unseen weave patterns.} Our network is trained on several typical patterns, similar to Jin et al.~\shortcite{Jin:2022:inverse}. Regarding the other weave patterns, the network needs retraining.}

\section{Conclusion}
\label{sec:conclusion}

In this paper, we presented a woven fabric parameter estimation pipeline using a captured reflection-transmission photo pair. The key component of the pipeline is our new fabric BSDF using an azimuth-invariant phase function to fit the multiple scattering of real back-lit fabrics better. Our full BSDF model has several components to match real woven fabric reflection and transmission. Our inverse framework allows a close match of both reflection and transmission to the input images.
Our lightweight capture can achieve high-fidelity recovery of woven fabrics at a distant view. However, we could further model yarn-level details and imperfections, and consider extensions to other types of fabrics, like knitted fabrics.



\begin{acks}
We thank the reviewers for the valuable comments. This work has been partially supported by the National Science and Technology Major Project under grant No.  2022ZD0116305 and National Natural Science Foundation of China under grant No. 62172220.
\end{acks}


\bibliographystyle{ACM-Reference-Format}
\bibliography{paper}



\begin{figure*}[tb]
\centering
\includegraphics[width = \linewidth]{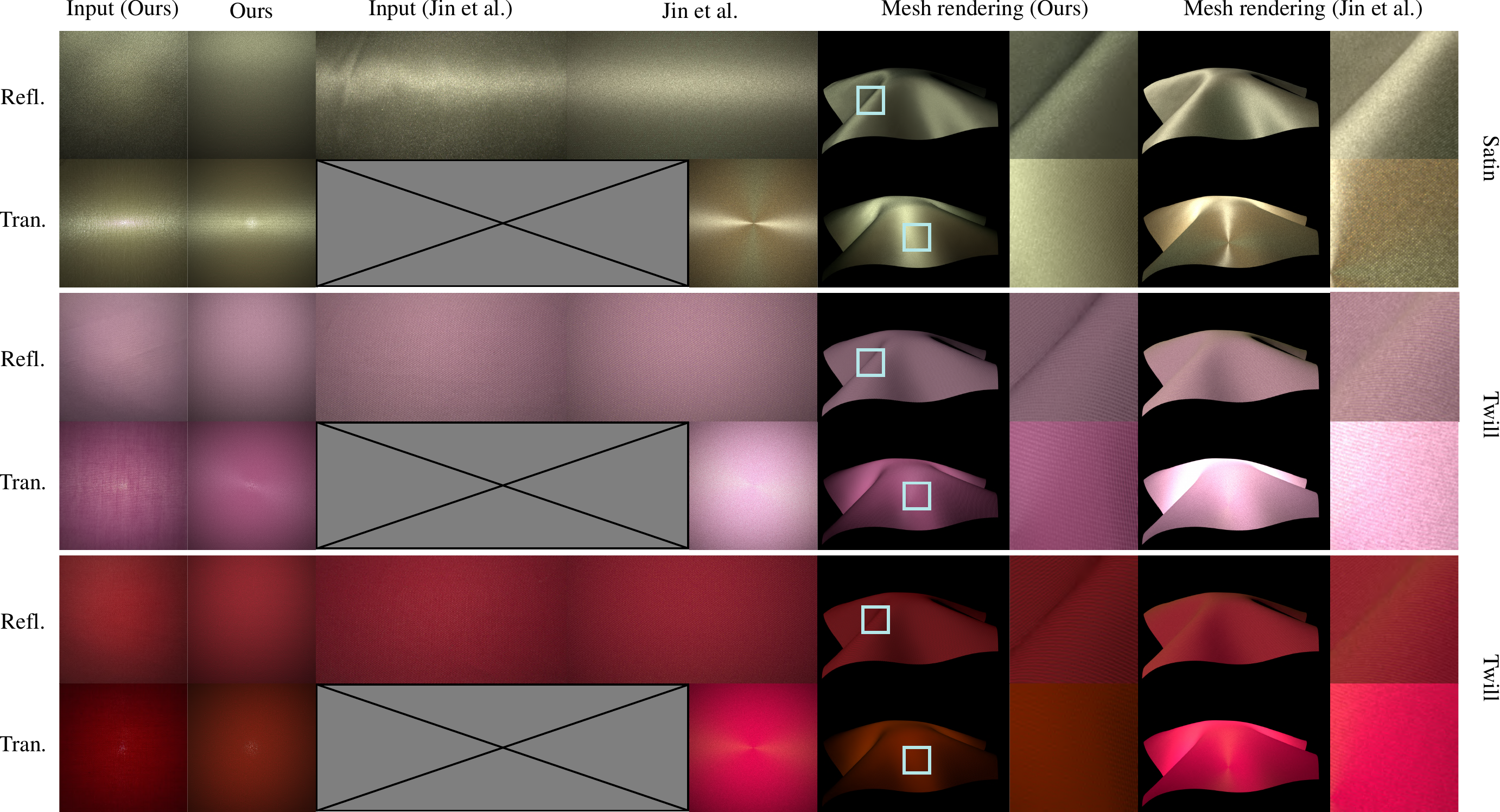}
\caption{Given an input image captured with our measurement configuration, our inverse model is able to produce closely matching results. The rendered results on the draped cloth mesh also show a natural appearance. \revise{Note that the shadowing-masking effects are included for the mesh renderings.}}
\label{fig:real}
\end{figure*}

\begin{figure}[tb]
\centering
\includegraphics[width = 0.75\linewidth]{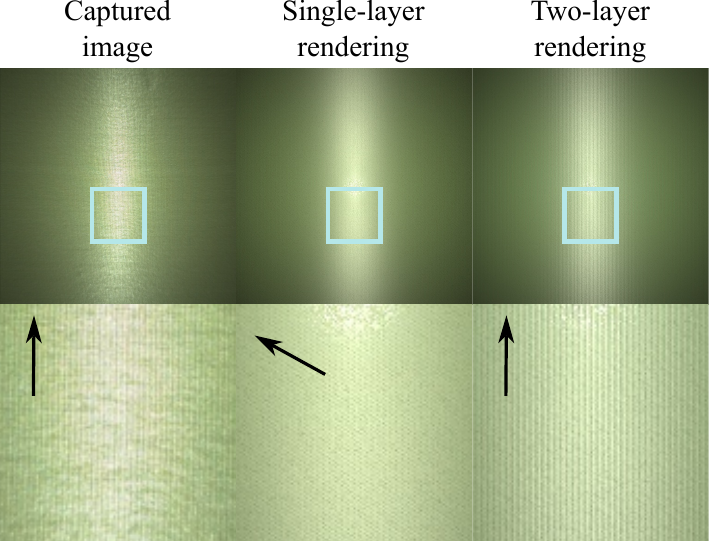}
\caption*{place holder}
\end{figure}

\begin{figure}[tb]
\centering
\includegraphics[width = 0.75\linewidth]{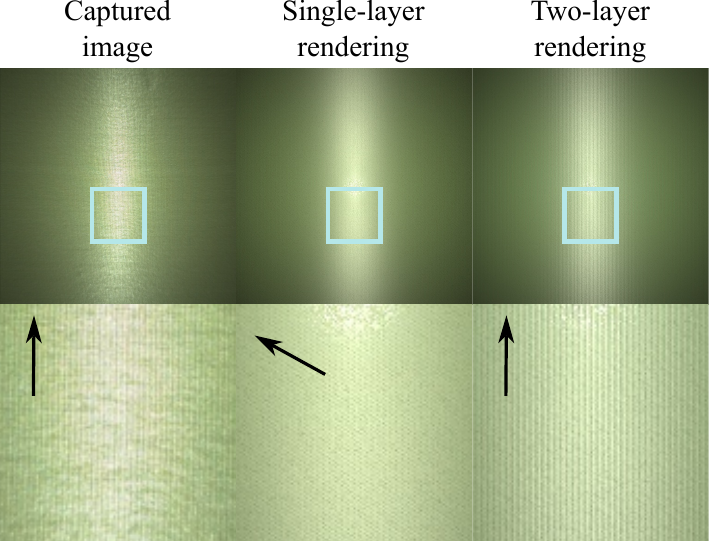}
\caption{The impact of the two-layer shading model. By comparing with the single-layer shading model, we find that our two-layer model can reproduce the continuous vertical yarns shown in the captured image. \revise{In contrast, the single-layer model shows oblique stripes (as the arrow indicates), which is inconsistent with the captured image.}}
\label{fig:ablation_twolayer}
\end{figure}

\begin{figure}[tb]
\centering
\includegraphics[width = 0.8\linewidth]{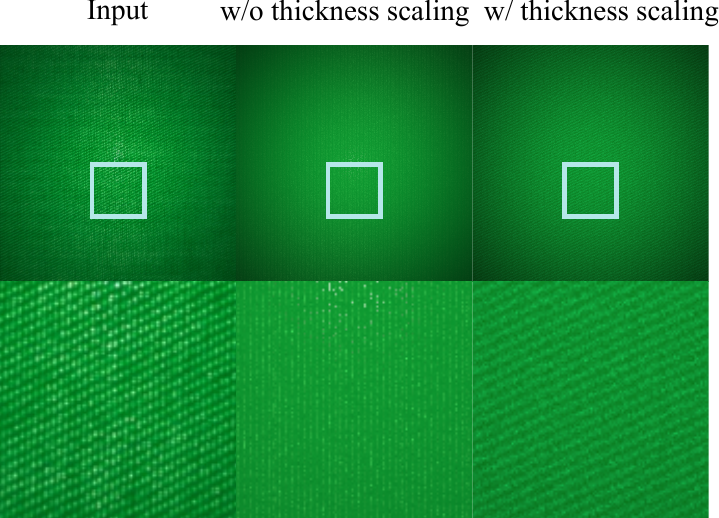}
\caption{Comparison between the rendered results with and without our tension-aware thickness scaling. With the tension-aware thickness scaling, the overall diagonal structure can be represented, which is missing in a constant thickness. }
\label{fig:ablation_thickscale}
\end{figure}

\begin{figure*}[tb]
\centering
\includegraphics[width = 0.95\linewidth]{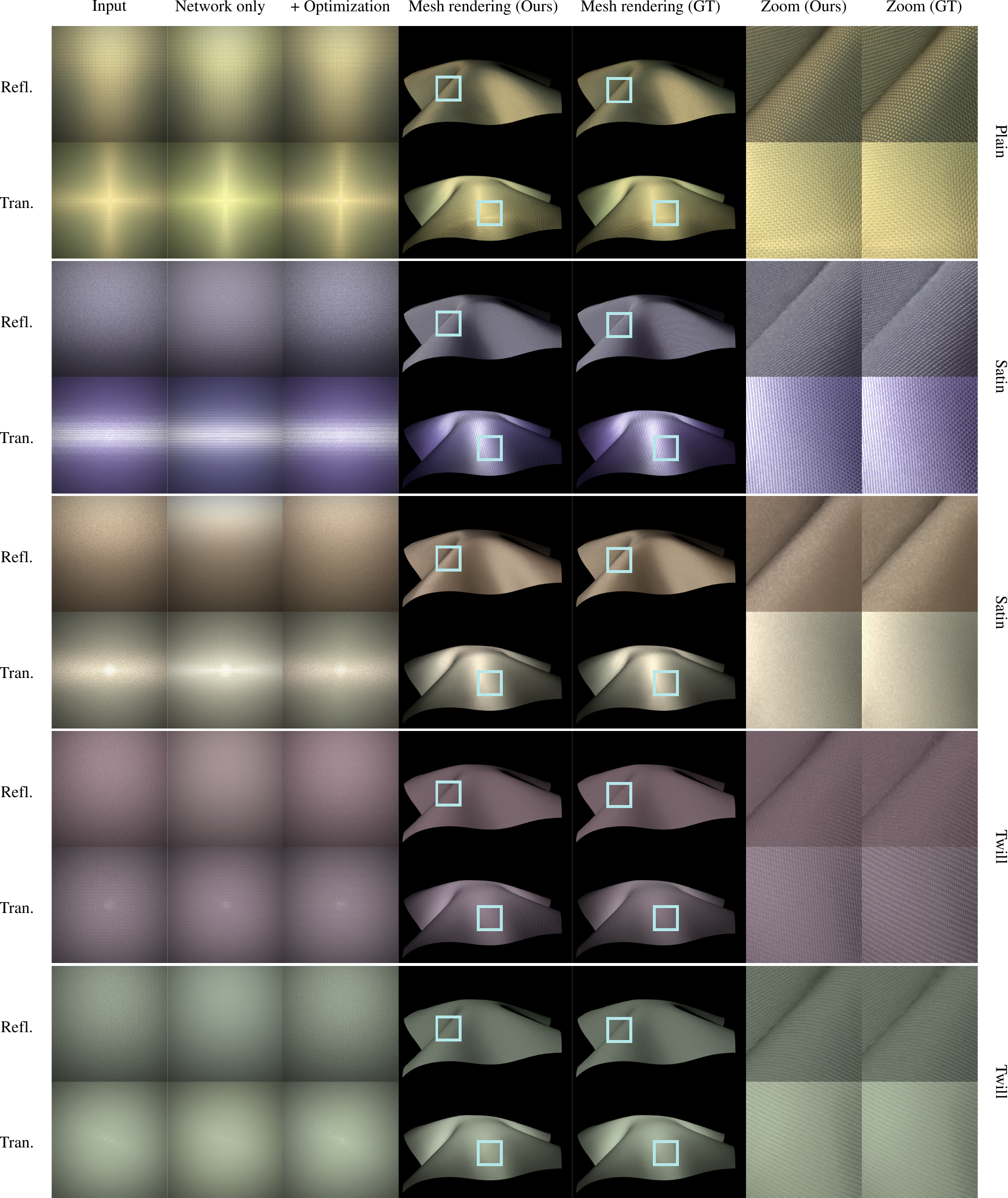}
\caption{Given synthetic input images, our neural network estimation can predict parameters that approach the appearance of the input. Using the optimization further improves the accuracy. Our results on the draped cloth mesh match the ground truth closely. \revise{Note that the shadowing-masking effects are included for the mesh renderings.} }
\label{fig:synthetic}
\end{figure*}

 


\end{document}


\author{Yingjie Tang}
\orcid{0009-0002-5633-0079}
\affiliation{
    \institution{Nankai University}
    \country{China}
}
\email{lxtyin.ac@gmail.com}

\author{Zixuan Li}
\orcid{0009-0004-2424-9529}
\affiliation{
    \institution{Nankai University}
    \country{China}
}
\email{zixuan.li_2001@outlook.com}

\author{Milo\v{s} Ha\v{s}an}
\orcid{0000-0003-3808-6092}
\affiliation{
    \institution{Adobe Research}
    \country{USA}
}
\email{milos.hasan@gmail.com}

\author{Jian Yang}
\orcid{0000-0003-4800-832X}
\affiliation{
    \institution{Nanjing University of Science and Technology}
    \country{China}
}
\email{csjyang@njust.edu.cn}

\author{Beibei Wang}
\orcid{0000-0001-8943-8364}
\authornote{Corresponding author.}
\affiliation{
    \institution{School of Intelligence Science and Technology, Nanjing University}
    \country{China}
}
\email{beibei.wang@nju.edu.cn}


\title{Supplementary materials: Woven Fabric Capture with a Reflection-Transmission Photo Pair}

\maketitle

\begin{table}[!t]
	\renewcommand{\arraystretch}{1.1}
	\caption{\label{tab:notations} Notation used for our BSDF model.}
\begin{small}
  \begin{tabular}{|l|c|l|}\hline
		$\win / \wout $ & incident / outgoing direction \\
		$\wout' $ & rotated outgoing direction \\
		$\omega_\mathrm{m} $ & macroscopic surface normal \\
		$\omega_\mathrm{n} $ & yarn normal \\
		$\omega_\mathrm{t} $ & yarn orientation \\
		$\omega_h $ & half vector between $\win$ and $\wout$ \\
  	$\omega_h' $ & half vector between $\win$ and $\wout'$ \\
		$D(\omega)$ &directional distribution of SGGX microflakes \\
		$G(\win, \wout)$ &shadowing-masking term for single scattering \\
		$D_m(\omega)$ &directional distribution of ASGGX microflakes \\
		$G_m(\win, \wout)$ &shadowing-masking term for multiple scattering\\
		$\sigma(\omega)  $ &projected area of microflakes\\
		$\rho$   & microflake density\\
		$T^{s,m}$ & layer thickness for single/multiple scattering \\
		$k_s^{s,m}$ & specular albedo for single/multiple scattering \\
		$\alpha^{s,m}$ & microflake roughness for SGGX/ASGGX \\
		$f_\mathrm{s}(\win, \wout)$ & SpongeCake single scattering BSDF \\
		$f_\mathrm{m}(\win, \wout)$ & Our multiple scattering BSDF \\
		$f_d^{r,t}(\win, \wout)$ & Our diffuse term for reflection and transmission \\
		$f^{\delta}(\win, \wout)$ & Delta transmission \\
  
				\hline
		\end{tabular}

\end{small}
\end{table}

\begin{table}
    \centering
    \setlength{\tabcolsep}{10pt}
    \caption{Distributions used to sample the parameter space of our model. The third column notes whether the parameter has separate versions for weft and warp. $\mathcal{U}(x, y)$ represents a continuous uniform distribution in the interval $(x, y)$. $\mathcal{V}(X)$ is a discrete uniform random variable on a finite set $X$. Yarn density is defined in yarns per inch, and converted internally to actual yarn size. Note that we didn't keep the multiple scattering albedo $k_s^m$ as a parameter directly, but use a weight term $w_m$ to control it, where $k_s^m = (k_s^s) ^{\frac{1.0}{w_m}}$.}
    \begin{tabular}{ccc} 
    \toprule
        Parameter & Sampling Function  & weft / warp \\
      \midrule
        yarn pattern & $ \mathrm{W} = \mathcal{V}(\{0, 1, 2, 3, 4\})$ & No \\
        yarn density & $y = \mathcal{U}(45, 335)$ & Yes\\
        roughness & $\alpha^{s,m} = \mathcal{U}(0.1, 1)^2$ & Yes \\
        thickness & $T^{s,m} = \mathcal{U}(0.1, 5)$ & Yes \\
        diffuse albedo & $k_\mathrm{d}^{r,t} = \mathcal{U}(0, 1)$ & No \\
        specular albedo & $k_\mathrm{s}^s = \mathcal{U}(0, 1)$ & Yes\\
        multiple weight & $w_\mathrm{m} = \mathcal{U}(0.1, 2)$ & No\\
        blending weight & $w = \mathcal{U}(0, 1)$ & No \\
        height field scaling & $\beta = \mathcal{U}(0.1, 2)$  & Yes\\
        gap scaling & $\mathrm{\xi} = \mathcal{U}(0.1, 1)$ & Yes \\
      \bottomrule
    \end{tabular}
    \label{tab:sample}
\end{table}

 \section{Our forward model}
\paragraph{Multiple scattering.} Our multiple scattering term is defined as:
\begin{eqnarray} 
    f_\mathrm{m}(\win, \wout) &= & \frac{k_s^m \ D_m(\omega_h') \ G_m(\win,\wout)} {2 \cti \cdot \cto}, \\
    G_m(\win,\wout) &= & \frac{1 - e^{-T^m \rho (\Lambda(\win) + \Lambda(\wout))}}{\Lambda(\win) + \Lambda(\wout)}, \\
    D_m(\omega) &= & \frac{1}{\pi \alpha^m q^2}, \mbox{ where }
    q = \omega^\top S^{-1} \omega,
    \label{eq:bsdf-flake}
\end{eqnarray}
where $\omega_i$ and $\omega_o$ represent the incident and outgoing directions respectively, $\omega_h'$ is the half-vector in our ASGGX by rotating $\omega_i$ and $\omega_o$ to the same plane. $S$ is a symmetric, positive definite $3 \times 3$ matrix defined in SGGX, $T^m$ and $\alpha^m$ is the thickness and roughness for the multiple scattering. The $G$ and $D$ terms are identical as the SpongeCake model~\cite{Wang:2021:Sponge}, except using the modified parameters.

\begin{figure}[tb]
\centering
\includegraphics[width = 0.8\linewidth]{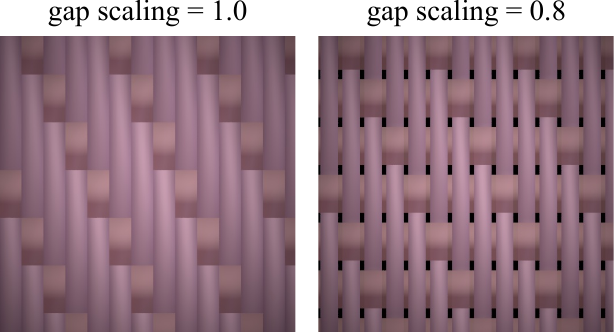}
\caption{The renderings with different gap scaling factors.}
\label{fig:gapscaling}
\end{figure}

\paragraph{Gap scaling} We scale the yarn width by a gap scaling factor to express the gaps between the weft and warp yarns, as shown in Fig. ~\ref{fig:gapscaling}. We use the same delta transmission as Zhu et al.~\shortcite{Zhu:2023:cloth} when light paths traverse the gaps:
\begin{equation} 
    f^{\delta}(\win ,\wout) = \frac{\delta(\win+ \wout)} {\left \langle \win \cdot \omega_m \right \rangle}.
\end{equation}

 \section{Discrete parameters optimization}
 
During optimization, we treat six parameters
as discrete: the yarn density of the weft and warp, the gap scaling of the weft and warp, and the twist angle for weft and warp. This is because we do not currently implement their gradients. With some effort, these gradients could be added, but our discrete solution gives good results. We randomly perturb these parameters every five iterations and accept the perturbation if it results in a lower loss than before. The yarn densities are perturbed as follows: +/- 10 yarns per inch before 100 iterations; +/- 5 yarns per inch from 100 to 150 iterations, and +/- 2 yarns per inch from 150 to 300 iterations. For the twist angles, it is perturbed as follows: +/- 0.5 degrees before 100 iterations; +/- 0.25 degrees from 100 to 150 iterations, and +/- 0.1 degree from 150 to 300 iterations. For the gap scalings, it is perturbed as follows: +/- 0.05 before 100 iterations; +/- 0.025 from
100 to 150 iterations, and +/- 0.01 from 150 to 300 iterations.

\section{More results}


\begin{table}
    \centering
    \caption{\revise{Comparison between the estimated and the ground-truth parameters (roughness and thickness for both warp and weft) on the synthetic data. The materials are shown in Fig. 16 (main paper).}}

    \begin{tabular}{l|c*{4}{p{0.05\textwidth}<{\centering}}} 
    
    \toprule

\multicolumn{2}{l}{\multirow{2}{*}{Sample}}
& \multicolumn{2}{c}{Roughness}
& \multicolumn{2}{c}{Thickness} \\
                        \cmidrule(lr){3-4} \cmidrule(lr){5-6}
\multicolumn{1}{l}{} & & warp & weft & warp & weft \\ \midrule

\rowcolor{mygray}
& GT & 0.26 & 0.26 & 1.23 & 1.23 \\
\rowcolor{mygray}
\multirow{-2}{*}{yellow plain}
& predicted	& 0.22 & 0.20 & 0.99 & 0.83 \\ \midrule

& GT & 0.22 & 0.90 & 1.12 & 3.75 \\
\multirow{-2}{*}{bule satin}
& predicted	& 0.28 & 0.80 & 1.27 & 3.30 \\ \midrule

\rowcolor{mygray}
& GT & 0.34 & 0.80 & 1.20 & 2.81 \\
\rowcolor{mygray}
\multirow{-2}{*}{brown satin}
& predicted	& 0.43 & 0.48 & 1.28 & 3.12 \\ \midrule

& GT & 0.87 & 0.93 & 3.98 & 1.08 \\
\multirow{-2}{*}{pink twill}
& predicted	& 0.81 & 0.91 & 2.46 & 1.82 \\ \midrule

\rowcolor{mygray}
& GT & 0.50 & 0.71 & 2.46 & 2.16 \\
\rowcolor{mygray}
\multirow{-2}{*}{green twill}
& predicted & 0.47 & 0.79 & 2.75 & 1.87 \\


    \bottomrule
    \end{tabular}
    \label{tab:syn_parameters}
\end{table}


\begin{figure*}[tb]
\centering
\includegraphics[width = \linewidth]{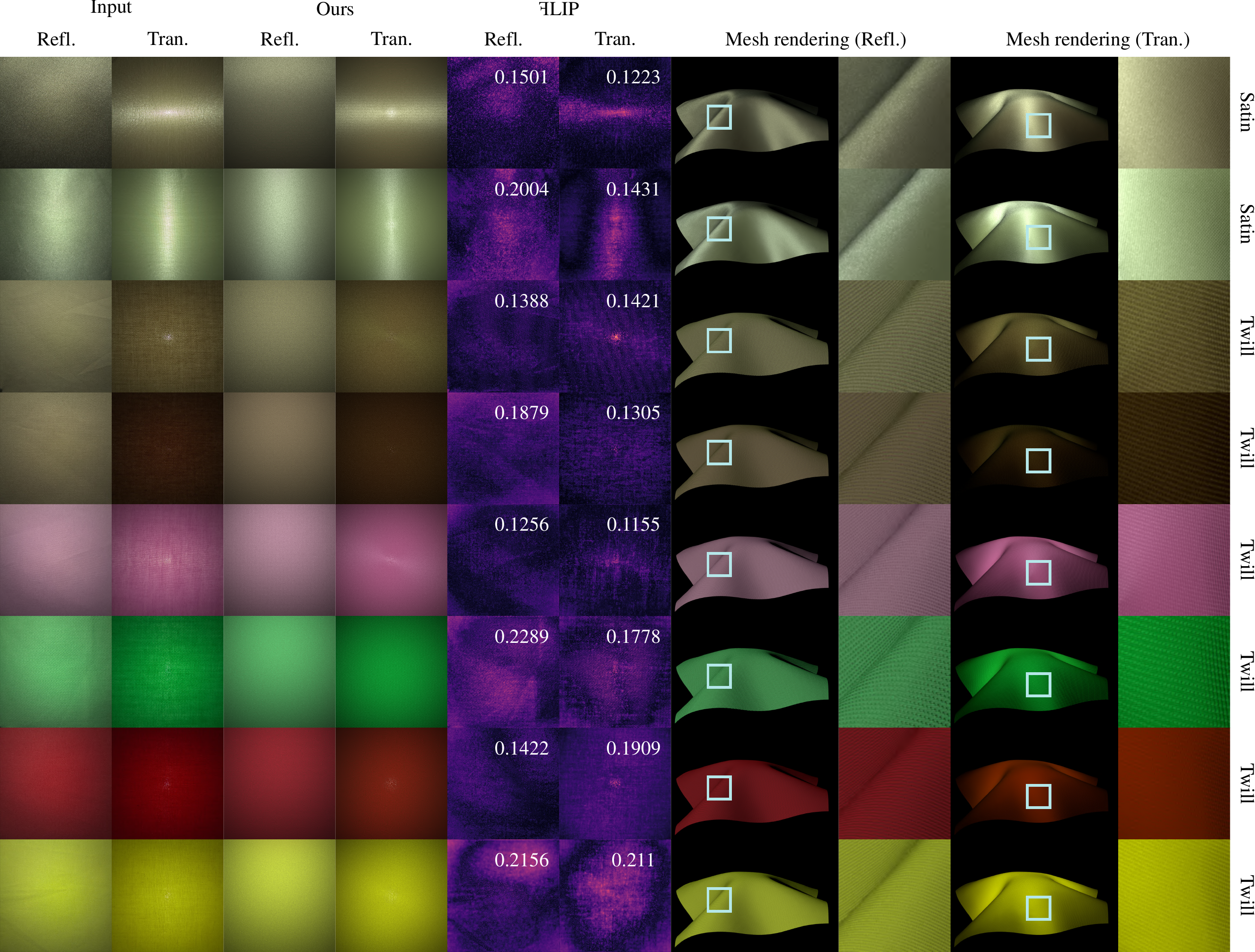}
\caption{
\revise{More results on the real captured data. The differences between the rendered and captured images are shown in the middle.}}
\label{fig:real}
\end{figure*}

\paragraph{Real data}
\revise{In Fig.~\ref{fig:real}, we provide more results on the real data, and report the {\FLIP} error between the captured and the recovered images. The renderings of our reconstructed parameters can match the captured images.}
\revise{In Fig.~\ref{fig:novelview}, we further validate our method by capturing the fabric samples from a novel view (by rotating them) and rendering them under the same configuration. By comparison, the rendering results of our estimated fabrics can match the captured fabrics at the novel view. Note that, in the transmission rendering, we include the out-of-focus effect by projecting the point light to the rendered image and generating a Gaussian around the projected center. }

\paragraph{Synthetic data}
\revise{We validate our method on synthetic data by comparing the estimated parameters (roughness and thickness for both warp and weft) to the ground-truth parameters. As shown in Table~\ref{tab:syn_parameters}, our recovered parameters are close to the ground truth.}

\paragraph{Impact of the loss function}
\revise{In Fig.~\ref{fig:ablation_loss}, we provide an ablation study for each loss term in the optimization step. We find that the pixel loss and the Gram matrix loss reduce the color bias, while the prior loss improves the robustness, especially for the gap scaling factor estimation.}



\begin{figure}[tb]
\centering
\includegraphics[width = \linewidth]{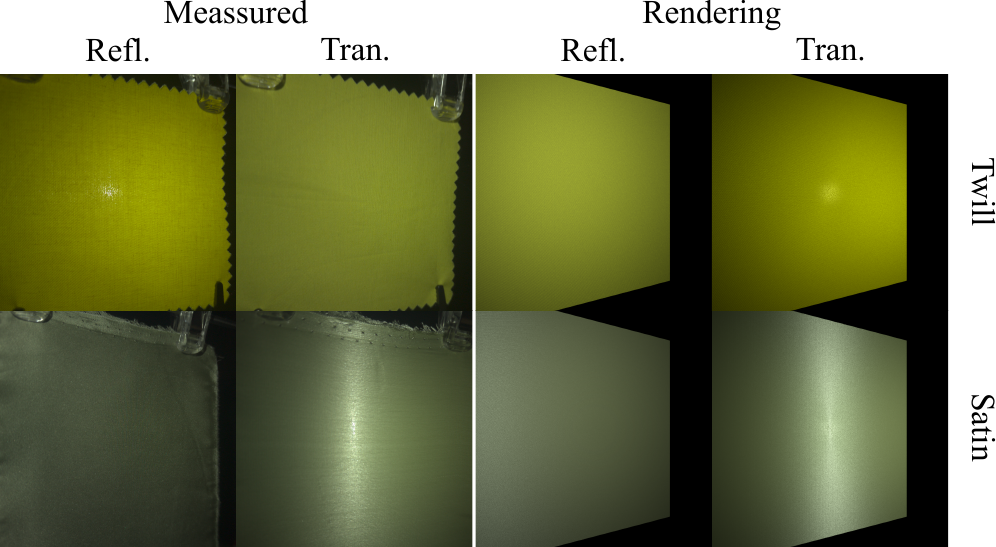}
\caption*{place holder}
\end{figure}

\begin{figure}[tb]
\centering
\includegraphics[width = 0.9\linewidth]{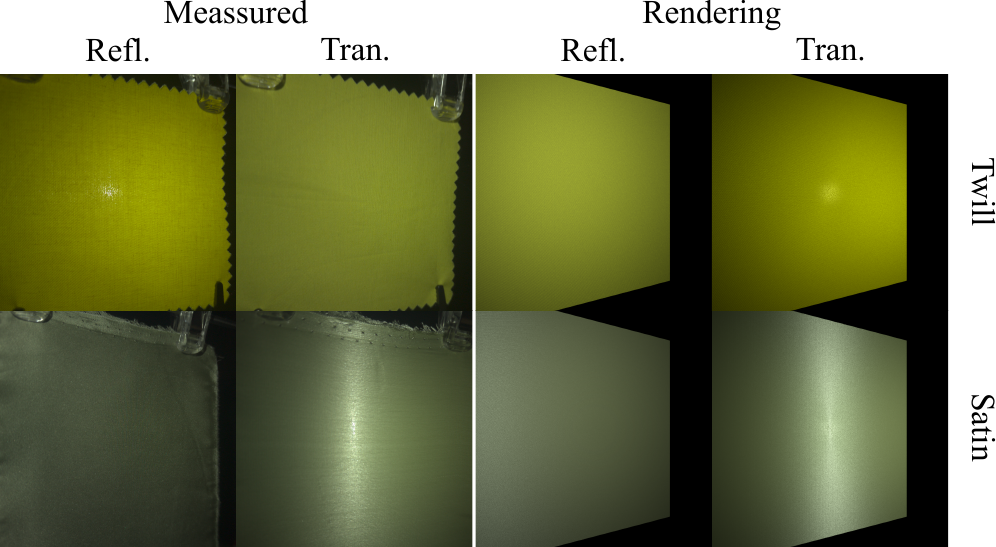}
\caption{
\revise{Novel view validation. We capture the fabrics from a novel view and compare it with our rendered result in the same configuration. Our results can match with real fabrics at novel view.}
}
\label{fig:novelview}
\end{figure}

\begin{figure}[tb]
\centering
\includegraphics[width = 0.9\linewidth]{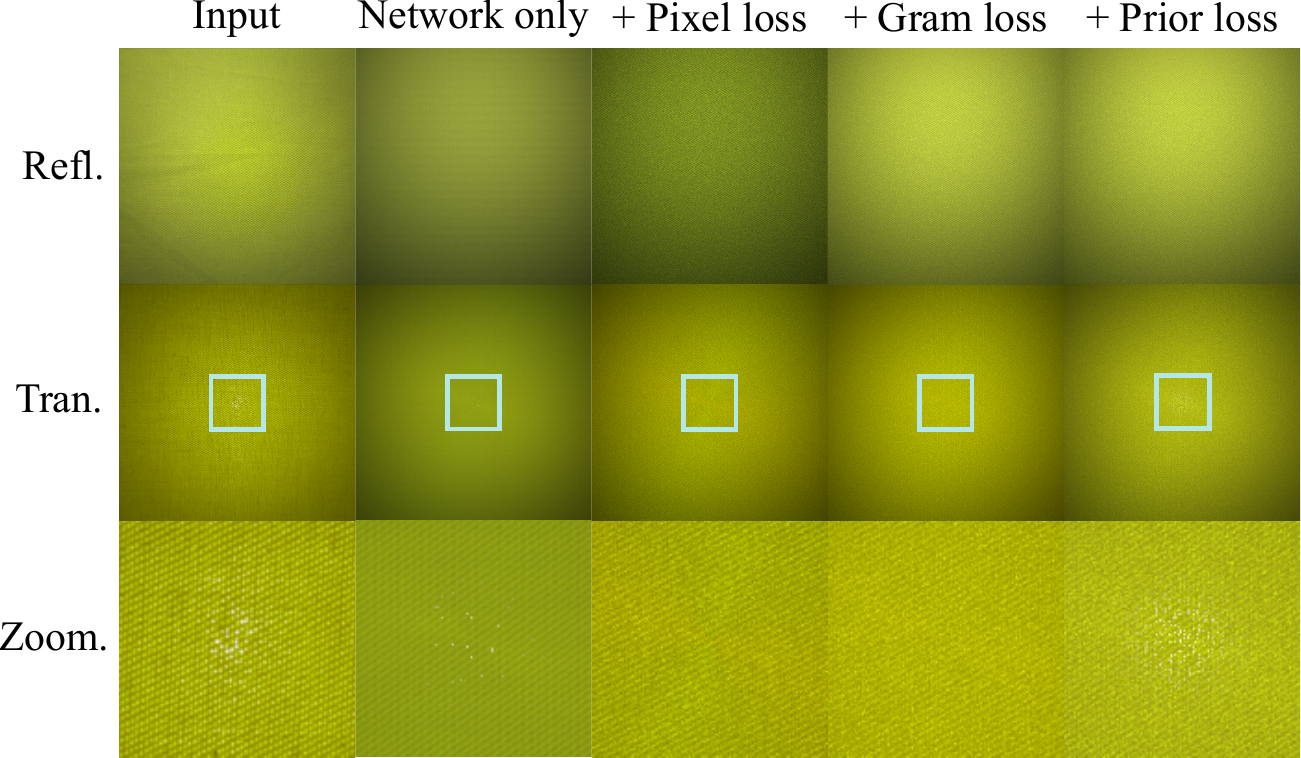}
\caption{
\revise{The impact of the loss terms. The pixel loss and the Gram matrix loss reduce color bias, and the prior loss helps find a suitable gap scaling factor, which contributes to the delta transmission in the center.}
}
\label{fig:ablation_loss}
\end{figure}






\bibliographystyle{ACM-Reference-Format}
\bibliography{paper}